\documentclass[10pt,twocolumn,letterpaper]{article}

\usepackage[pagenumbers]{wacv} 

\usepackage{graphicx}
\usepackage{amsmath}
\usepackage{amssymb}
\usepackage{booktabs}

\usepackage{float}
\usepackage{colortbl}
\usepackage{algorithm}
\usepackage[noend]{algpseudocode}
\usepackage{multirow} 
\usepackage{xcolor}

\usepackage[pagebackref,breaklinks,colorlinks]{hyperref}

\usepackage[capitalize]{cleveref}
\crefname{section}{Sec.}{Secs.}
\Crefname{section}{Section}{Sections}
\Crefname{table}{Table}{Tables}
\crefname{table}{Tab.}{Tabs.}

\begin{document}

\title{DiTAS: Quantizing Diffusion Transformers via Enhanced Activation Smoothing}
\author{Zhenyuan Dong\\
New York University\\
{\tt\small zd2362@nyu.edu}
\and
Sai Qian Zhang\\
New York University\\
{\tt\small sai.zhang@nyu.edu}
}
\maketitle

\begin{abstract}
Diffusion Transformers (DiTs) have recently attracted significant interest from both industry and academia due to their enhanced capabilities in visual generation, surpassing the performance of traditional diffusion models that employ U-Net. However, the improved performance of DiTs comes at the expense of higher parameter counts and implementation costs, which significantly limits their deployment on resource-constrained devices like mobile phones. We propose~\text{DiTAS}, a data-free post-training quantization (PTQ) method for efficient DiT inference. DiTAS relies on the proposed temporal-aggregated smoothing techniques to mitigate the impact of the channel-wise outliers within the input activations, leading to much lower quantization error under extremely low bitwidth. To further enhance the performance of the quantized DiT, we adopt the layer-wise grid search strategy to optimize the smoothing factor. Moreover, we integrate a training-free LoRA module for weight quantization, leveraging alternating optimization to minimize quantization errors without additional fine-tuning. Experimental results demonstrate that our approach enables 4-bit weight, 8-bit activation (W4A8) quantization for DiTs while maintaining comparable performance as the full-precision model. Code is available at \href{https://github.com/DZY122/DiTAS}{https://github.com/DZY122/DiTAS}.
\end{abstract}

\section{Introduction}
\label{sec:intro}

\begin{figure}[t]
  \centering
   \includegraphics[width=1.25\linewidth, trim={1.7cm 1.5cm 6cm 0.5cm}, clip]{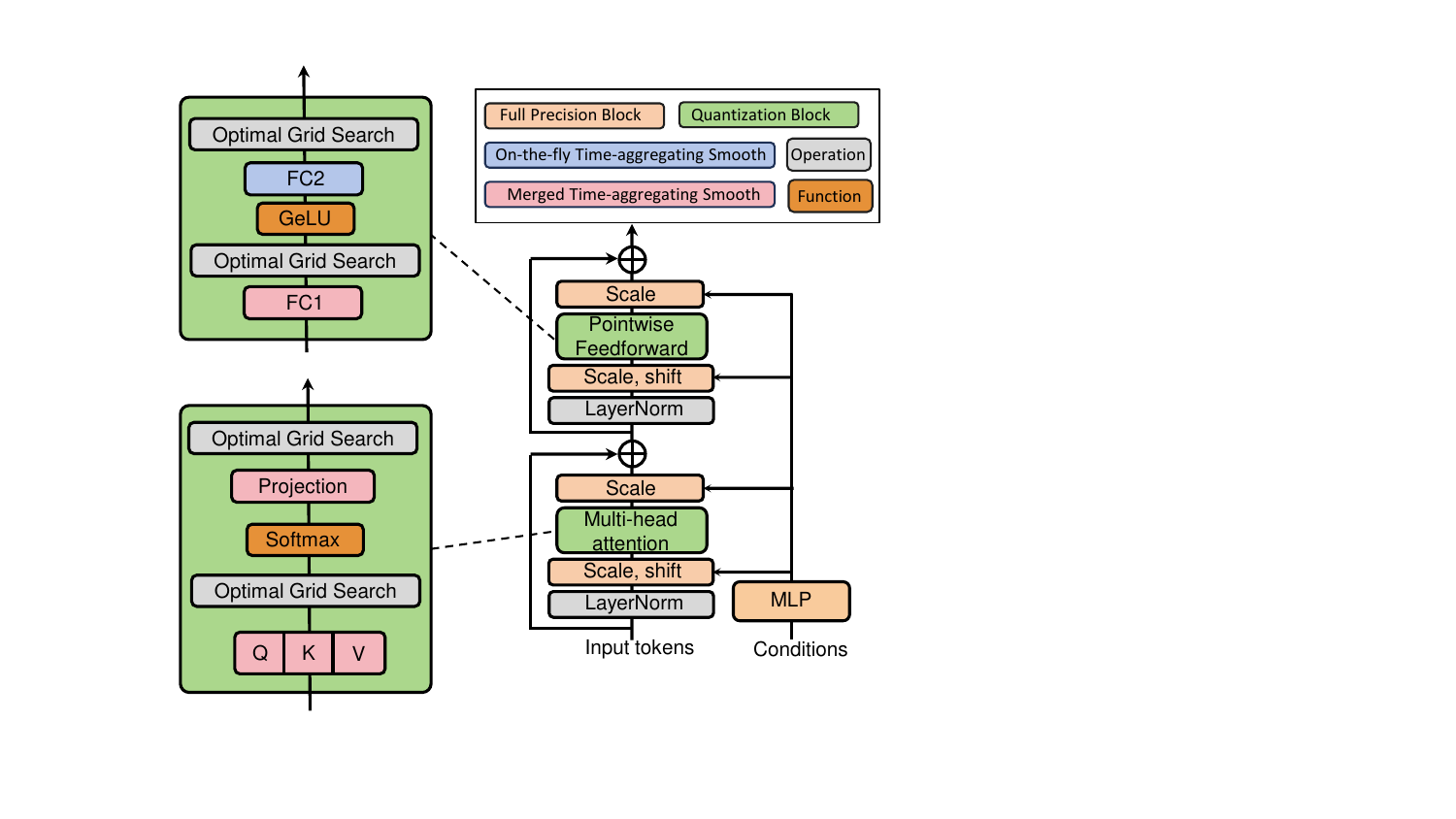}
   \caption{DiTAS architecture.}
   \label{fig:DiTAS_arch}
\end{figure}

\begin{figure*}[t]
\centering
\begin{minipage}[t]{0.34\linewidth}
\includegraphics[width=1\linewidth, trim={0cm 0cm 0cm 0cm}, clip]{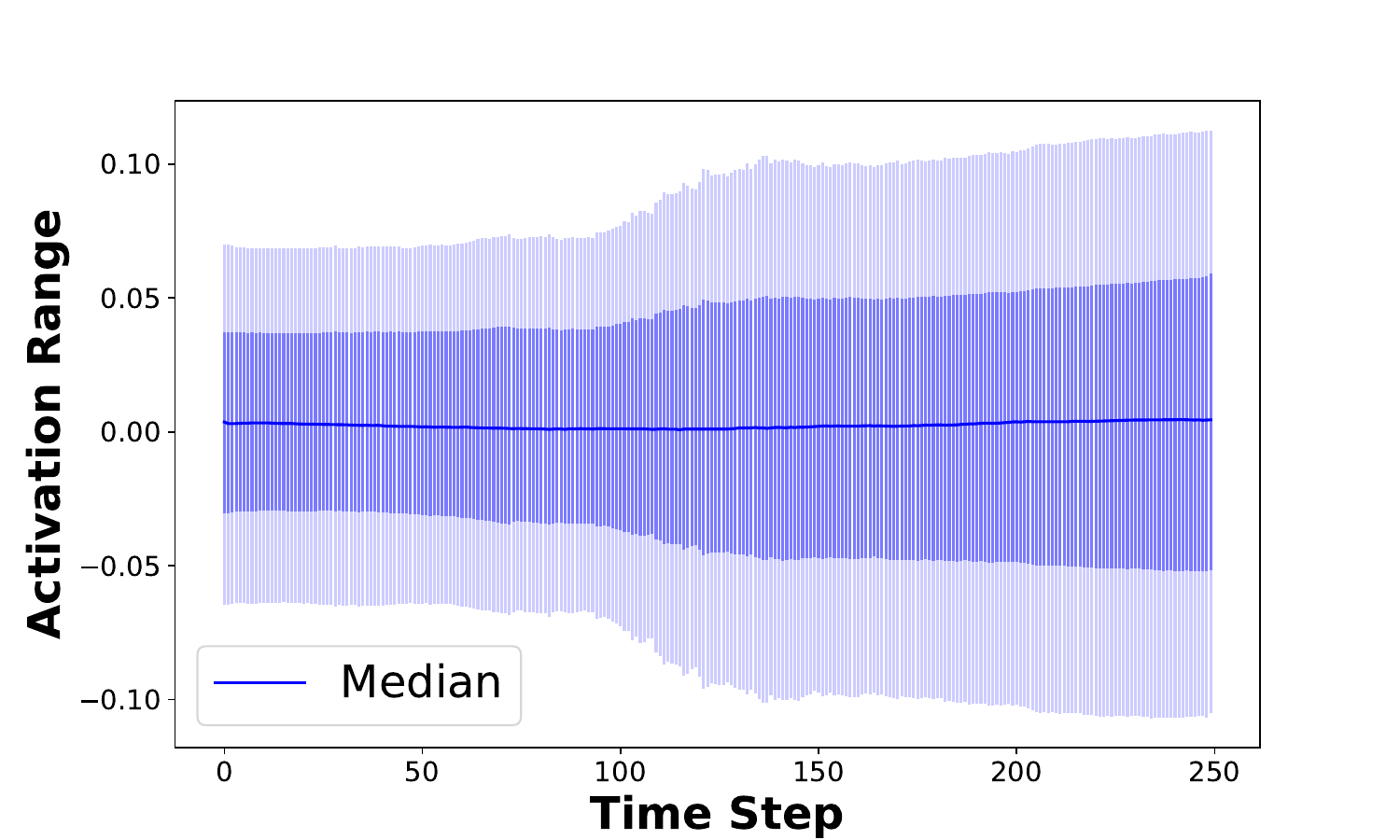}
\caption{Input activation range across different time steps. The dark blue segment shows the 95th percentile range, the light blue segment denotes the extreme values.}
\label{ACT}
\end{minipage}
\hfill
\begin{minipage}[t]{0.295\linewidth}
\includegraphics[width=1\linewidth]{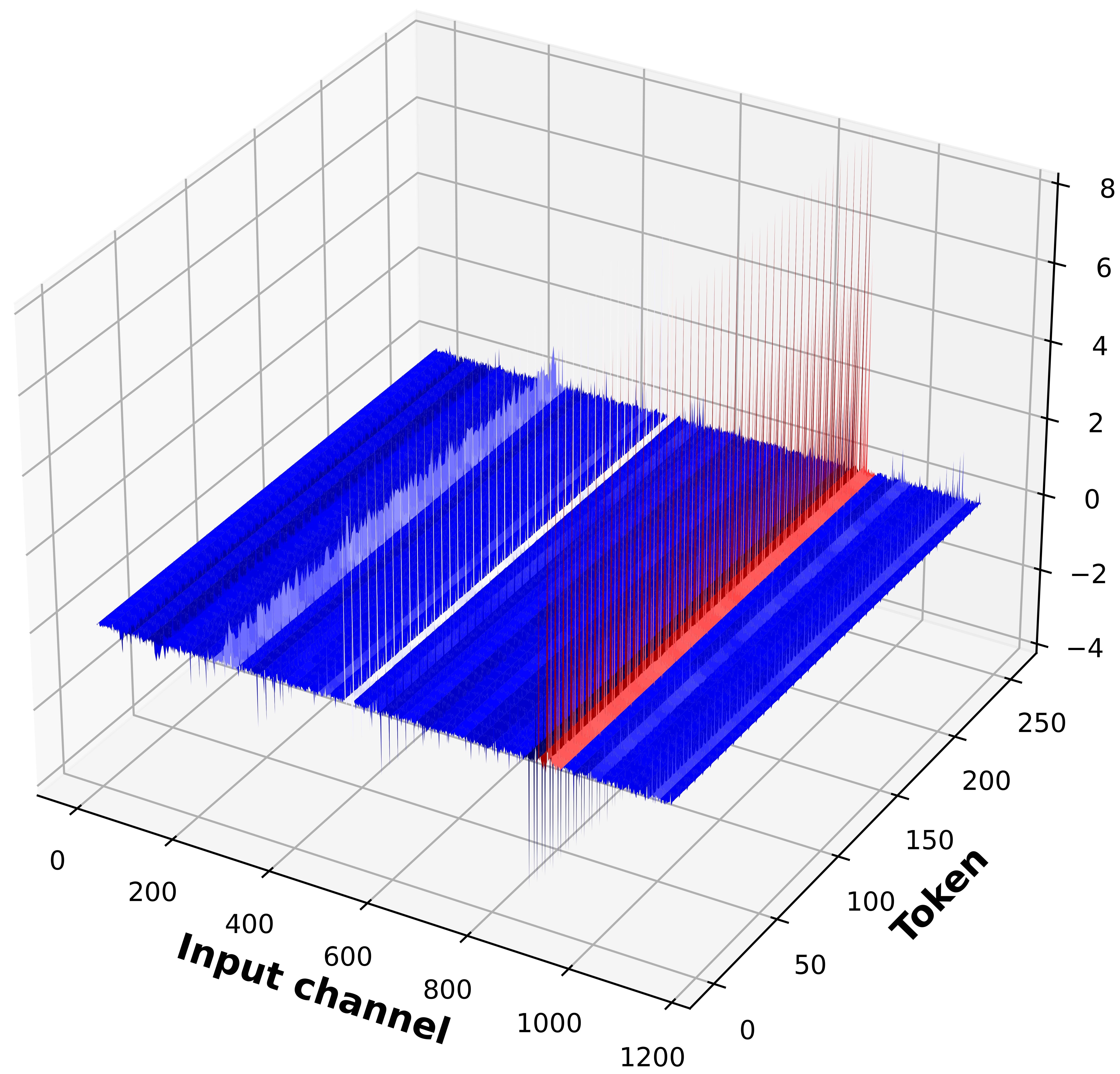}
\caption{Activation range before Temporal-aggregated Smoothing (TAS).}
\label{TCS1}
\end{minipage}
\hfill
\begin{minipage}[t]{0.295\linewidth}
\includegraphics[width=1\linewidth]{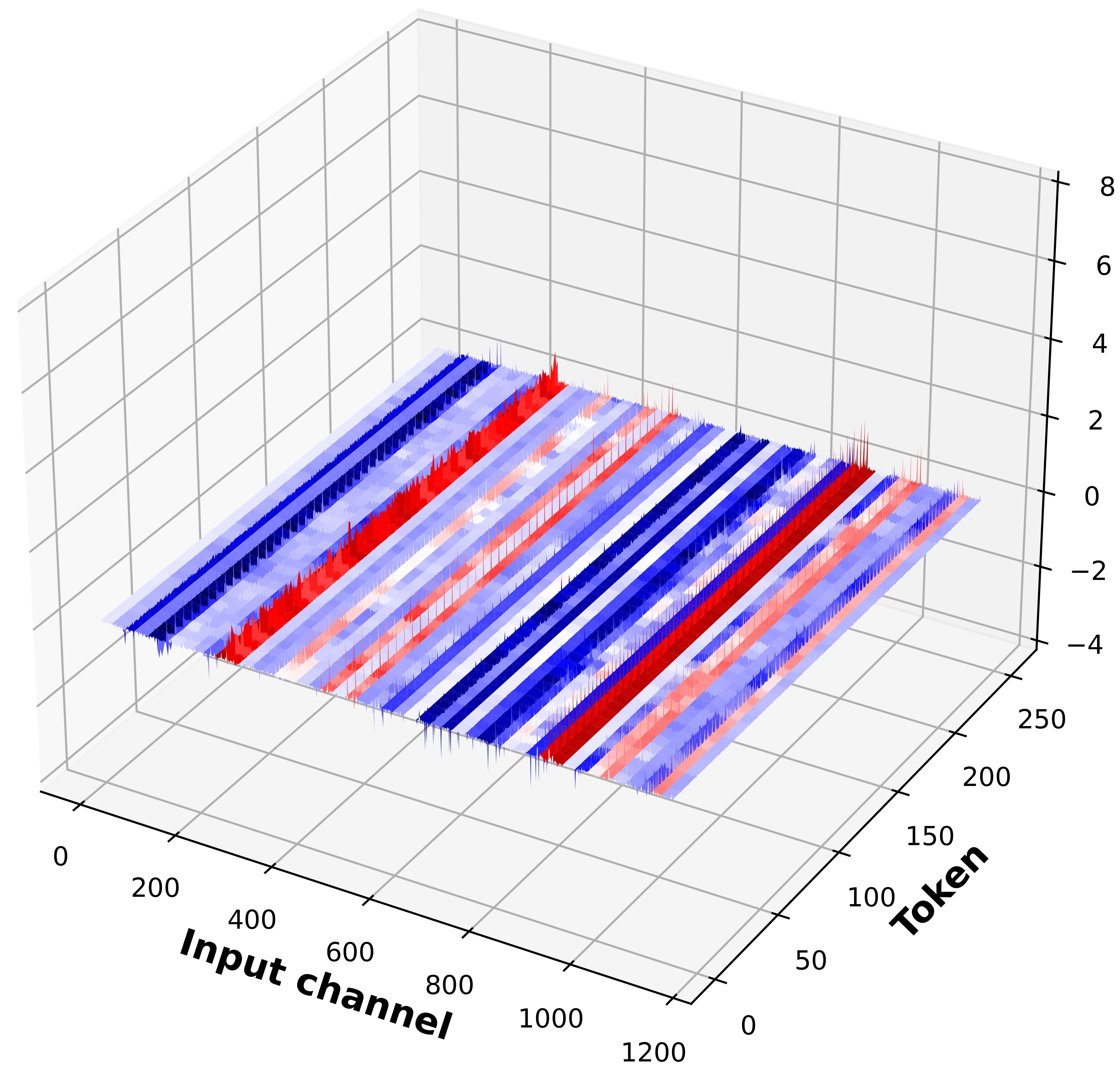}
\caption{Activation range after Temporal-aggregated Smoothing (TAS).}
\label{TCS2}
\end{minipage}
\end{figure*}

Diffusion transformers (DiTs)~\cite{peebles2023scalable} have gained significant attention due to their superior performance compared to traditional diffusion models (DMs) that utilize U-Net~\cite{rombach2022ldm} as the backbone deep neural network. Since their introduction, DiTs have been extensively researched and applied in both academic and industrial fields~\cite{peebles2023scalable, mo2024dit, feng2024latent, wu2024medsegdiff, gao2023masked}, with notable applications such as OpenAI's SoRA~\cite{Sora}. Recent studies have demonstrated their impressive generative capabilities across various modalities~\cite{gao2024lumina}. 

However, the iterative denoising steps and substantial computational requirements significantly slow down their execution. Although various methods have been proposed to reduce the thousands of iterative steps to just a few dozen, the large number of parameters and the complex network structure of DiT models still impose a considerable computational burden at each denoising step. This limitation hinders their practicality in resource-constrained environments.

Model quantization is widely acknowledged as an effective strategy for reducing memory and computational demands by compressing weights and activations into lower-bit representations. Among the various quantization techniques, Post-Training Quantization (PTQ) provides a training-free approach (or minimal training cost for calibration purposes~\cite{nagel2020ptq, li2021brecq, lin2021fq}) for rapid and efficient quantization. Compared to Quantization-Aware Training (QAT), which requires multiple rounds of fine-tuning, PTQ incurs significantly lower computational costs. This makes it an appealing option for quantizing large models like DiT. Existing PTQ methods for diffusion models (DMs)~\cite{li2023q, he2024efficientdm} primarily use fixed-point quantization (i.e., INT quantization); however, substantial quantization errors can occur at low precision, resulting in poor performance.


In this study, we introduce DiTAS, an efficient quantization method for low-precision DiT execution with minimal impact on the generated image quality. We observe considerable variance in activation distribution across time steps, often accompanied by the presence of channel-wise outliers. To mitigate these effects, we adopt~\textit{temporal-aggregated smoothing} (TAS). TAS involves channel-wise smoothing factor aggregates information about the magnitude of outliers from all time steps, which can effectively mitigate the visual quality degradation caused by the outliers. 

To further enhance the performance of quantized DiT, we employ a layer-wise grid search strategy to optimize the smoothing factor in TAS. TAS, along with the optimized factor, effectively eliminates outliers in the input activations, further enhancing DiT performance at extremely low precision levels. In addition to the efficient activation smoothing techniques, inspired by the concept of LoftQ~\cite{li2024loftq}, we integrate Low-Rank Adaptation (LoRA) modules~\cite{hu2021lora,han2024parameter} into quantized DiT weights and apply Alternating Optimization (AO) to boost performance. From Figure~\ref{fig:DiTAS_arch}, the DiTAS architecture is designed to operate TAS and grid search optimization layer by layer. For the QKV and FC1 layers in DiT blocks, we merge the smoothing factor of activation into the side MLP. And we merge the smoothing factor of Projection layer's activation into V's weight. Finally, we operate on-the-fly activation smoothing for FC2 layers. In summary, we have the following contributions:

\begin{itemize}
\item We introduce temporal-aggregated smoothing to minimize the impact of outliers within input activations performance of quantized DiT model. This can effectively mitigate the impact of activation outliers on the quantized DiT performance.
\item We propose a layer-wise grid search optimization strategy to fine-tune the smoothing factors for each input channel across time steps, aiming to better reduce the impact of outliers within the activations. Additionally, to address quantization errors in the weights, we introduce the LoRA module over the weights and employ AO to minimize the quantization error in the weights, allowing the adjusted weights to closely approximate the original values, thereby enhancing overall quantized DiT performance.

\item Extensive experiments on ImageNet at resolutions of $256 \times 256$ and $512 \times 512$ show that our DiTAS achieves state-of-the-art performance in DiT quantization at low precisions. Specifically, under W4A8 configurations with a Classifier-Free Guidance (cfg) score of 1.50, DiTAS achieves FID-10K scores of 9.05 for 50 sampling steps and 6.86 for 100 sampling steps, respectively, on ImageNet $256 \times 256$. Additionally, the W4A8 configurations of DiTAS (cfg=1.50) on ImageNet $512 \times 512$ achieves FID-10K scores of 17.92 for 50 sampling steps and 13.45 for 100 sampling steps, respectively.
\end{itemize}


\section{Related Work}
\label{sec:related-work}

\subsection{Diffusion Models Quantization}
\label{sec:bg:ptq1}

DMs have recently garnered significant attention for their remarkable ability to generate diverse photorealistic images. These models are parameterized Markov chains trained via variational inference to generate samples that match the data distribution over a finite duration. The large model size of DMs and the high implementation cost make them impractical for deployment on resource-limited devices, presenting a significant challenge for real-time applications on various mobile devices. Therefore, leveraging PTQ to compress models into smaller scale can effectively improve the efficiency. Unlike QAT, PTQ does not necessitate model training, or it incurs only minimal training cost for calibration purposes, making it highly computationally cost-effective. When calibrating generative models such as DMs, rather than utilizing the original training dataset, the calibration datasets can be generated using the full-precision model. This approach enables the calibration process to be implemented in a data-free manner. For example, Q-Diffusion~\cite{li2023q} adopts a reconstruction-based PTQ approach in DMs, while PTQD~\cite{he2024ptqd} integrates decomposed quantization errors into the random noise. QNCD~\cite{chu2024qncd} presents a unified quantization noise correction scheme designed to reduce quantization noise throughout the sampling process. EfficientDM~\cite{he2024efficientdm} enhances the performance of the quantized DM by fine-tuning the model using quantization-aware low-rank adapter.Some papers also explore the impact of varying time-steps on DM quantization. APQ-DM~\cite{wang2024towards} develops distribution-aware quantization functions for activation discretization at different timesteps and optimize the selection of timesteps for generating informative calibration images. TFMQ-DM~\cite{huang2024tfmq} introduces a temporal feature maintenance quantization framework based on a temporal information block that focuses specifically on the time-step t, enabling temporal information aware reconstruction.

\subsection{Diffusion Transformer Quantization}
\label{sec:bg:ptq2}

Given the growing popularity of DiT, recent research has also focused on quantizing DiT at low precision~\cite{wu2024ptq4dit,chen2024q,liu2024hq}. PTQ4DiT~\cite{wu2024ptq4dit} employs Channel-wise Salience Balancing (CSB) and Spearman’s \(\rho\)-guided Salience Calibration (SSC) to mitigate quantization errors in DiTs. Q-DiT~\cite{chen2024q} introduces a group size allocation algorithm for fine-grained quantization of both activations and weights in DiTs. In contrast, our approach differs from these existing DiT quantization methods. However, all previous approaches result in noticeable visual quality drop of the generated images at resolutions of $256 \times 256$ and $512 \times 512$. In contrast, DiTAS surpasses all previous approaches in terms of generated image quality, particularly under low-precision scenarios.

\section{Methodology}
\label{sec:methodology}
In this section, we describe the DiTAS in detail. We first introduce the DiT background in Section~\ref{sec:prelim}. Next we describe our TAS techniques in Section~\ref{sec:tcs-quant}, followed by the grid search optimization in Section~\ref{sec:grid_search}. We will finally introduce the advanced weight quantization in Section~\ref{sec:LoRA-quant}.


\subsection{Preliminaries}
\label{sec:prelim}
\paragraph{Diffusion Transformers.}
Diffusion Transformers (DiTs) is a new architecture for diffusion models, surpassing the performance of traditional diffusion models that employ U-Net. The architecture of a DiT block is depicted in Figure~\ref{fig:DiTAS_arch}. DiT is built upon transformer-based Diffusion Models (DDPMs)~\cite{ho2020denoising}. Both the training strategy and the inference process closely resemble those of traditional DDPMs. As a Markov chain, Gaussian diffusion models operate under the assumption of a forward noise process that gradually introduces noise to the real data $x_0$:
\begin{equation}
q\left(x_t \mid x_0\right)=\mathcal{N}\left(x_t ; \sqrt{\bar{\alpha}_t} x_0,\left(1-\bar{\alpha}_t\right) \mathbf{I}\right)
\end{equation}
where constants $\bar{\alpha}_t$ are hyperparameters, which can be chosen and fixed. With the parameterization, we have $x_t=\sqrt{\bar{\alpha}_t} x_0+\sqrt{1-\bar{\alpha}_t} \epsilon_t$, where $\epsilon_t \sim \mathcal{N}(0, \mathbf{I})$.
Diffusion models are trained to learn a Gaussian distribution: 
\begin{equation}
p_\theta\left(x_{t-1} \mid x_t\right)=\mathcal{N}\left(x_{t-1} ; \mu_\theta\left(x_t\right), \Sigma_\theta\left(x_t\right)\right)
\end{equation}
where $\epsilon_\theta$ and $\Sigma_\theta$ are the statistics prediction from transformer-based neural networks.

\paragraph{Asymmetric INT Quantization.}
Asymmetric quantization is a widely adopted method for quantizing deep neural networks. It involves mapping the weights or activations of a DNN from 32-bit floating-point numbers to a low precision data format (e.g., INT). Asymmetric quantization offers a dynamic mapping range compared to symmetric quantization, allowing for more flexible and accurate quantization models. Specifically, given an input \(\mathbf{x}\), its quantized version \(\hat{x}\) can be compuated using the following formula:
\begin{equation}
\label{eq:ptq}
\small
\widehat{\mathbf{x}}=q(\mathbf{x} ; s, z, b)=s\left[\operatorname{clamp}\left(\left\lfloor\frac{\mathbf{x}}{s}\right\rceil+z ; 0,2^b-1\right)-z\right]
\end{equation}
where $\lfloor \cdot \rceil$ is the $\mathrm{round}$ operation and $b$ is the bitwidth. $s$ and $z$ are the quantization scale and zero-point which are determined by the lower and upper bound of quantization thresholds. $clamp(x,min,max)$ generates a clipped version of input x by restricting it between $min$ and $max$.

\begin{algorithm}[t]
\caption{Grid Search Optimization}\label{Smooth}
\footnotesize
\begin{algorithmic}[1]
\Require Pretrained DiT model with $L$ linear layers; Total number of denoising time step $T$; Generative calibration dataset $D$.
\For{all $l=1,2,\dots, L$} 
    \State Collect the weight as $\mathbf{W}$
    \State Compute $ \max_{1 \leq t \leq T, 1 \leq b \leq B, 1 \leq l \leq L} (\left| \mathbf{X}_{btlc} \right| )$ from $D$ as $a_{c}$ 
    \State Compute $ \max_{1 \leq n \leq N} \left(\left|\mathbf{W}_{nc}\right|\right)$ from $D$ as $b_{c}$
    \State Let the final TAS factor  as $\mathbf{s}_{l}$ 
    \State Let  $\mathbf{s} = 0$ 
    \State Let $\mathcal{L}_{min}= \infty$
    \For{all $m=0, 1,2,\dots, 20$} 
        \State $\alpha = 0.05 \times m$
        \State $\mathbf{s} = [\frac{a_{1}^\alpha}{b_{1}^{(1-\alpha)}}, \frac{a_{2}^\alpha}{b_{2}^{(1-\alpha)}}, \ldots, \frac{a_{C}^\alpha}{b_{C}^{(1-\alpha)}}]$ 
        \State Let $\mathcal{L}=0$
        \For{all $t=1,2,\dots, T$} 
            \State  Collect input activation $\mathbf{X}_{t}$ of FP32 DiT from dataset $D$.
            \State  Collect output activation $\mathbf{Y}_{t}$ of FP32 DiT from dataset $D$.
            \State $\mathbf{Y}_{q}=Q(\mathbf{X}_{t} \operatorname{diag}(\mathbf{s})^{-1}) Q( \operatorname{diag}(\mathbf{s})\mathbf{W}) + bias$
            \;
            \State Compute $\mathcal{L}_t=\left\|\mathbf{Y}_{q}-\mathbf{Y}_{t}\right\|^2$
            \
            \State Compute $\mathcal{L}=\mathcal{L} + \mathcal{L}_t$
        \EndFor
        \If{$\mathcal{L}_{min} > \mathcal{L}$}
            \State $\mathbf{s}_{l} = \mathbf{s}$
            \State $\mathcal{L}_{min}= \mathcal{L}$
        \EndIf
    \EndFor
\EndFor
\State \Return optimized $\mathbf{s}_{l}$ for each layer.
\end{algorithmic}
\end{algorithm}


\begin{table}[t]
\centering
\begin{tabular}{cc}
\toprule
Method                    & FID$\downarrow$     \\ \midrule
Select time-step 1 to operate SmoothQuant                              & 261.92             \\
Select time-step 25 to operate SmoothQuant   & 109.48      \\
Select time-step 50 to operate SmoothQuant  & 151.82      \\
\rowcolor{blue!10}Temporal-aggregated Smoothing (TAS)                & \textbf{22.31} \\ \bottomrule
\end{tabular}
\caption{Smoothing methods comparison under W4A8 configuration on ImageNet $256 \times 256$ (cfg=1.5, 50 steps)} 
\label{tab:TAS_Smooth}
\end{table}

\subsection{Temporal-aggregated Smoothing}
\label{sec:tcs-quant}
To understand the distribution of input activations within DiT, we conduct an experiment where we collect the input activations of a DiT block across the entire denoising steps. The resulting histogram is illustrated in Figure~\ref{ACT}. Subsequently, Figure~\ref{TCS1} shows the distribution of the activation matrix at a specific time step across each input channel. Specifically, we profile the input activations of the 28th DiT block's first feed-forward layer by conducting forward propagation with a randomly chosen class label. During this process, we record the maximum and minimum activation values for each input channel. We make the following observations: First, there is a significant variation in the activation range across different time steps. Additionally, within the same time step, the activation range also varies significantly across different channels due to the presence of outliers. 
As indicated by the previous works~\cite{wu2024ptq4dit, he2024efficientdm, xiao2023smoothquant}, these outliers will cause substantial quantization errors, which can further degrade the performance of DiT under low quantization precision.

To address this, SmoothQuant~\cite{xiao2023smoothquant} introduces a per-channel smoothing factor $\mathbf{s} \in \mathbb{R}^{c_{\text{in}}}$ to alleviate the impact of activation outliers in large language models (LLMs). However, due to the unique nature of DiT, the distribution of input activations varies significantly across different timesteps $T$, as illustrated in Figure~\ref{ACT}. From Table~\ref{tab:TAS_Smooth}, we can find out directly applying the SmoothQuant method to DiT by selecting calibration data from a single timestep could potentially degrade the performance of the quantized DiT.

To address this issue, we propose the \textit{Temporal-aggregated Smoothing (TAS)} by introducing a channel-wise smoothing factor that aggregates information about the magnitude of outliers across all time steps, effectively managing both temporal variability and outliers. Specifically, the scaling factor can be computed as follows:

\begin{equation}\label{eqn:smooth}
    \mathbf{s}_{c} = \frac{ \max_{1 \leq t \leq T, 1 \leq b \leq B, 1 \leq l \leq L} (\left| \mathbf{X}_{btlc} \right| )^\alpha}{\max_{1 \leq n \leq N} \left(\left|\mathbf{W}_{nc}\right|\right)^{1-\alpha}}
\end{equation}

where $\mathbf{s}_{c}$ is the scaling factor for $c$-th input channel. $\mathbf{X}$ is a four-dimensional tensor with a shape of $B\times T\times L\times C$, where $B, T, L, C$ represent the batch size of the calibration data, total time steps, token length and input channels, respectively. 
$\mathbf{W}$ is a two-dimension matrix with a shape of $N\times C$, where $N$ and $C$ denote the number of output and input channels, respectively. 
The hyperparameter $\alpha$ determines the extent to which we aim to shift the impact of outliers from activations to weights. $\alpha$ is assigned a value of 0.5 to achieve this balance.  As indicated by equation~\ref{eqn:smooth}, the value of the scaling factor $\mathbf{s}_{c}$ is derived by taking into account both the input and weight distributions across different timesteps. The output $Y_{t}$ of each linear layer at time step t can be described as follows:
\begin{equation}
\label{eq:fine-tune}
\mathbf{Y}_t=Q(\mathbf{X}_t \operatorname{diag}(\mathbf{s})^{-1}) Q( \operatorname{diag}(\mathbf{s})\mathbf{W})
\end{equation}
Where $\mathbf{s} = [\mathbf{s}_1, \mathbf{s}_2, \ldots, \mathbf{s}_C]$ is a vector composed of elements \(\mathbf{s}_j\), each representing a distinct channel. $\text{diag}(\textbf{s})$ denotes the diagonal matrix whose diagonal elements consists of the elements of $\mathbf{s}$. $Q(.)$ is the quantization function described in Section~\ref{sec:prelim}, $\mathbf{W}$ is the pre-trained weight with full precision, and $\mathbf{X}_t$ is the input activation at $t$-th time step. 

\subsection{Grid Search Optimization}
\label{sec:grid_search}
The temporal-aggregated smoothing factor in our paper serves to dynamically alleviate the impact of channel-wise activation outliers across time steps. Our proposed grid search strategy is to better balance the extent to which we aim to shift the impact of outliers from activations to weights. To identify the most effective TAS factor, we can make the parameter \( s \) learnable, allowing it to better adapt to the current data distribution.
\begin{equation}
    \label{eqn:scale-loss}
    \mathbf{s}^* = \mathop{\arg\min}_{\mathbf{s}} \mathcal{L}(\mathbf{s}) 
\end{equation}
Instead of using backpropagation to train the scaling factor $s$, we utilize grid search optimization to find the optimal $s$ that minimizes the difference between the outputs of the linear layers generated using the quantized versions of weights and inputs compared to their FP32 counterparts. This approach ensures robust optimization by accounting for the temporal dynamics of the model's performance, while also eliminating the need for the costly backpropagation operations. The loss function can be defined as follows:
\begin{equation}
\small
\mathcal{L}(\mathbf{s}) = \sum_{t=1}^{T} \left\|Q(\mathbf{X}_t \operatorname{diag}(\mathbf{s})^{-1}) Q( \operatorname{diag}(\mathbf{s})\mathbf{W}) - \mathbf{X}_t \mathbf{W}\right\|^2
\end{equation}

Moreover, given that the scaling factor s is a function of $\alpha$, as depicted in equation~\ref{eqn:smooth}. We can reformulate equation~\ref{eqn:scale-loss} as follows:
\begin{equation}
\label{eq:scale_search_formula}
    \alpha^*=\mathop{\arg\min}_{\alpha}\mathcal{L}(\frac{a_{1}^\alpha}{b_{1}^{(1-\alpha)}}; \frac{a_{2}^\alpha}{b_{2}^{(1-\alpha)}};  \ldots; \frac{a_{C}^\alpha}{b_{C}^{(1-\alpha)}})
\end{equation}

where $a_{c} = \max_{1 \leq t \leq T, 1 \leq b \leq B, 1 \leq l \leq L} (\left| \mathbf{X}_{btlc} \right| )$ and $b_{c}= \max_{1 \leq n \leq N} \left(\left|\mathbf{W}_{nc}\right|\right)$. The detailed grid searching algorithm is described in Algorithm~\ref{Smooth}.




\begin{figure}[t]
\centering
\hfill
\begin{minipage}[ht]{0.49\linewidth}
\includegraphics[width=1\linewidth]{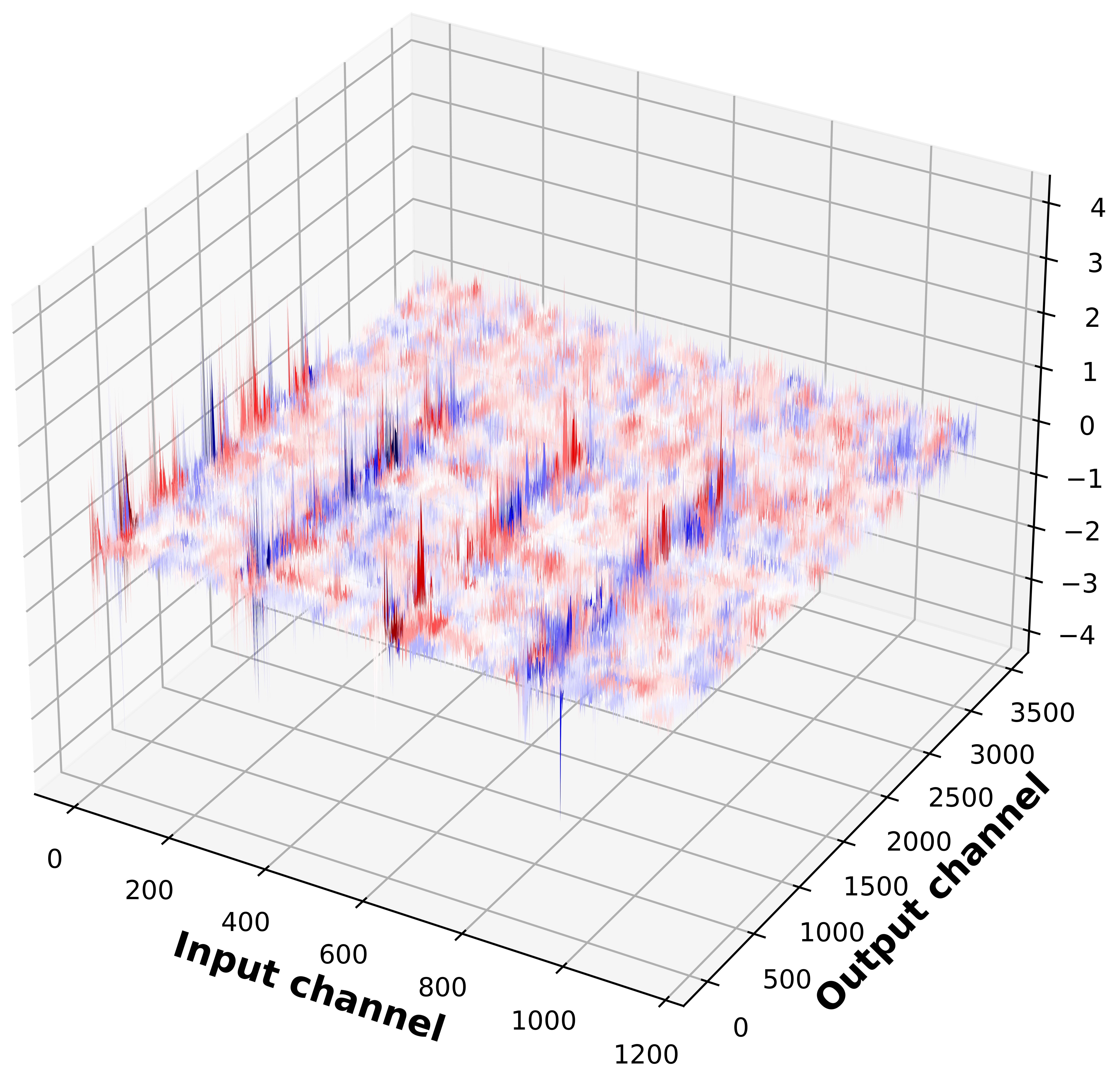}
\caption{Weight with outliers across input channels in 7th DiT Block's QKV layer.}
\label{weight_outlier_1}
\end{minipage}
\hfill
\begin{minipage}[ht]{0.49\linewidth}
\includegraphics[width=1\linewidth]{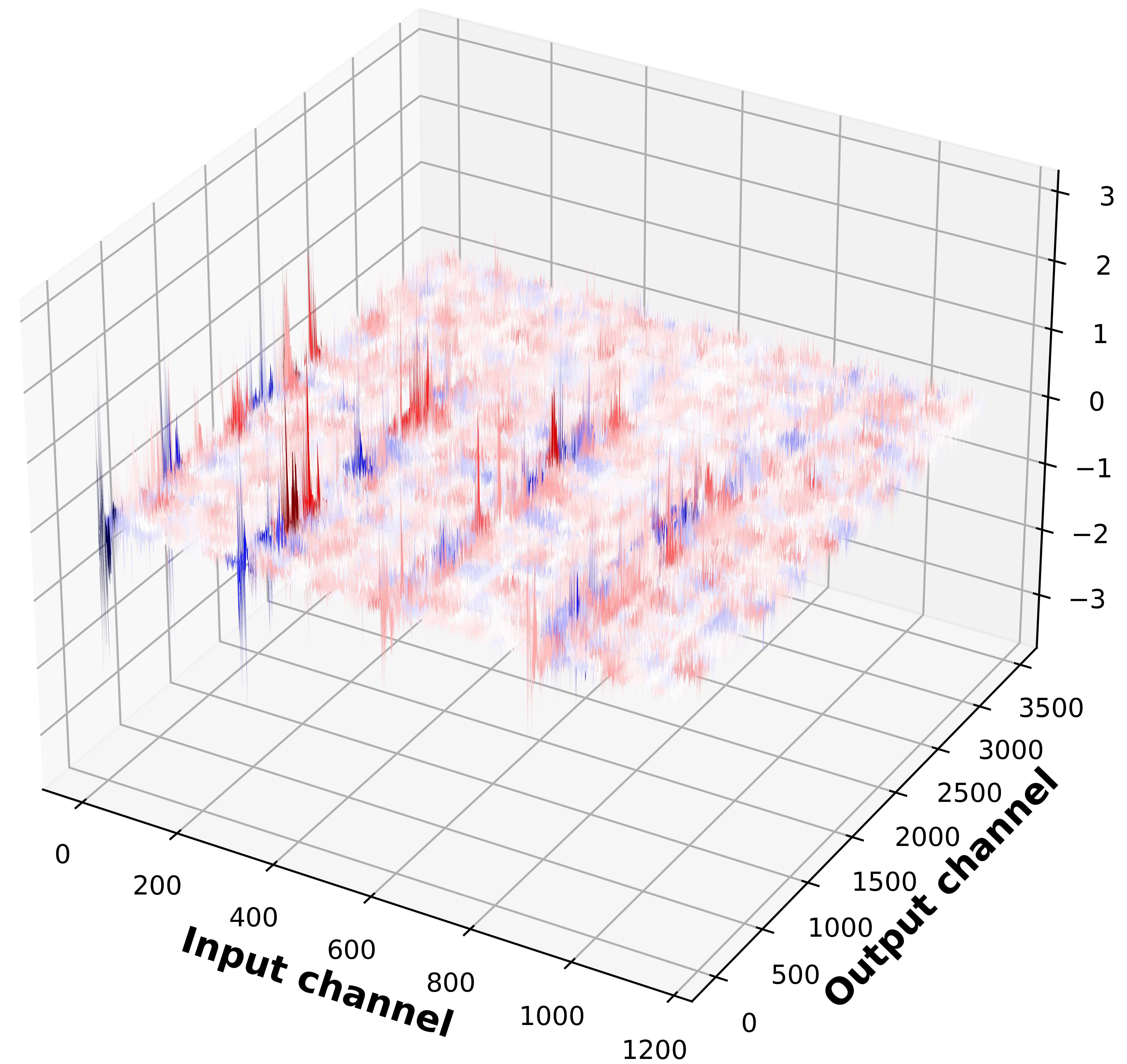}
\caption{Weight with outliers across input channels in 5th DiT Block's QKV layer.}
\label{weight_outlier}
\end{minipage}
\end{figure}

\begin{figure*}[t]
\centering
\hfill
\begin{minipage}[ht]{0.49\linewidth}
\includegraphics[width=0.99\linewidth, trim={0cm 18.5cm 0cm 0cm}, clip]{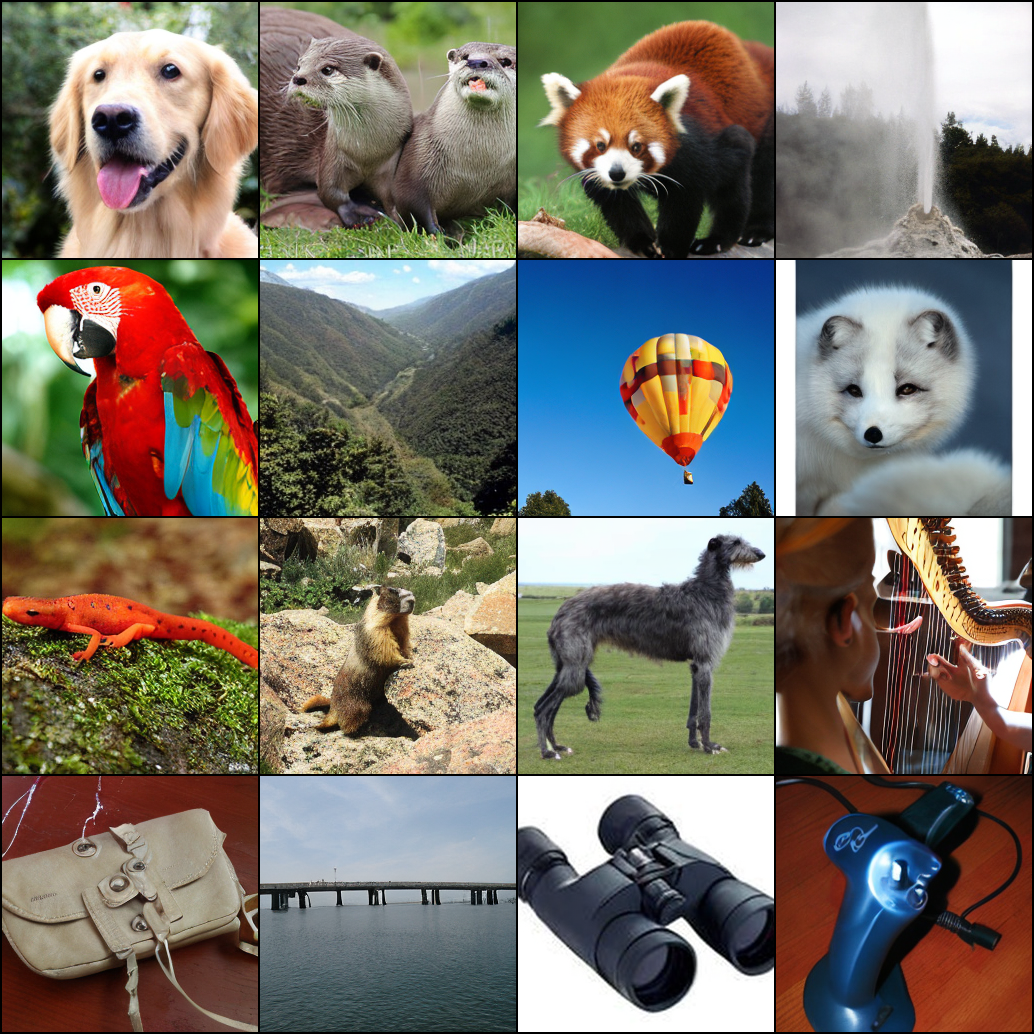}
\caption{Samples generated by W8A8 DiTAS model by 100 steps on ImageNet $256 \times 256$ (cfg=4.0).}
\label{W8A8}
\end{minipage}
\hfill
\begin{minipage}[ht]{0.49\linewidth}
\includegraphics[width=0.99\linewidth, trim={0cm 18.5cm 0cm 0cm}, clip]{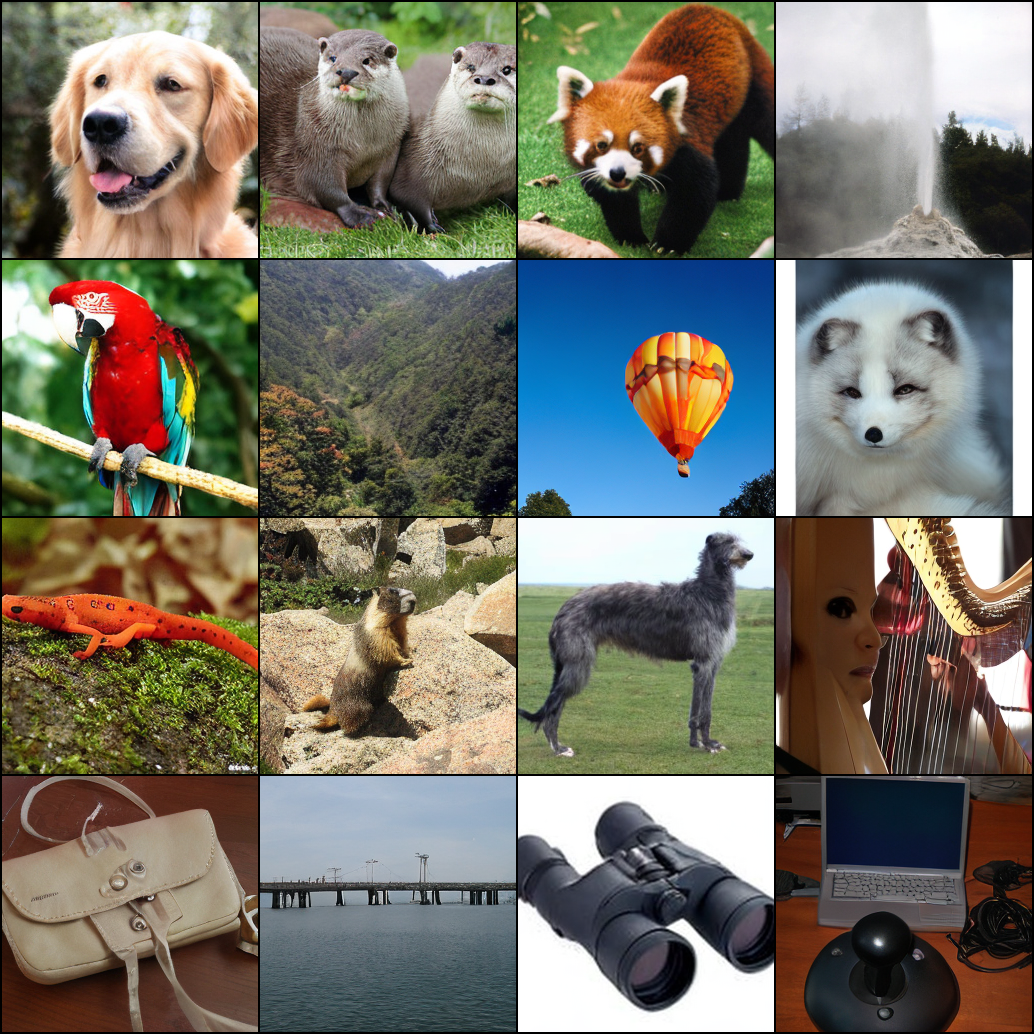}
\caption{Samples generated by W4A8 DiTAS model by 100 steps  on ImageNet $256 \times 256$ (cfg=4.0).}
\label{W4A8}
\end{minipage}
\end{figure*}

\subsection{Activation and Weight Quantization}
\label{sec:LoRA-quant}
After implementing the activation smoothing techniques outlined in Section~\ref{sec:tcs-quant} and Section~\ref{sec:grid_search}, we then describe the quantization approach applied to both activations and weights. Given the dynamic nature of activations, quantization needs to be applied adaptively. To minimize extra computational overhead, we utilize the asymmetric INT quantization method detailed in Section~\ref{sec:prelim}. Next, we depict the advanced weight quantization techniques in detail.

\paragraph{Fine-Grained Weight Quantization}
The weight matrix is quantized at the channel level, with each input channel being quantized separately. This approach is based on our observation that the weight elements within the input channel dimension are much smaller in magnitude compared to those in the output channel as shown in figure~\ref{weight_outlier_1} and figure~\ref{weight_outlier}. Consequently, quantizing over the input channel results in a smaller quantization error.


\paragraph{Alternating Optimization for LoRA Integration}
To compensate for the weight quantization error, we use the AO method to obtain a LoRA module in a data-free manner. This approach aims to significantly efficiently improve the performance of the DiTAS. The LoRA module does not require subsequent fine-tuning, specifically targeting the linear layers within the quantized DiT. Fine-tuning transformers requires substantial computational resources. So that this approach circumvents the memory and computational costs associated with subsequent LoRA fine-tuning by directly optimizing the quantized weights to numerically approximate the FP32 weights. Specifically, after introducing the LoRA module, the output $Y_{t}$ of each linear layer at time step t can be described as follows:
\begin{equation}
\label{eq:fine-tune}
\small
\mathbf{Y}_t=Q(\mathbf{X}_t^*) (\text{diag}(\mathbf{s})Q(\mathbf{W}))+Q(\mathbf{X}_t^*) (\text{diag}(\mathbf{s}) \mathbf{A}) \mathbf{B}^{\top}
\end{equation}
where $Q$ is the quantization function, $\mathbf{X}_t$ is the FP32 input activation at time step t, and $\mathbf{W}$ is the FP32 weight. $\mathbf{X}_t^*=\mathbf{X}_t \operatorname{diag}(\mathbf{s})^{-1}$. $\mathbf{B} \in \mathbb{R}^{N \times r}$ and $\mathbf{A} \in \mathbb{R}^{C \times r}$ are the full precision learnable matrices with $r \ll$ $\min \left(c_{in}, c_{out}\right)$, where $r$, $C$, and $N$ represent the rank of the LoRA module, the number of input channels, and output channels of the weight matrix $W$, respectively.

We employ the AO approach to have an optimized LoRA module without fine-tuning, where we adopt singular value decomposition (SVD) to obtain a low-rank approximation of the quantization error, and gain a newly quantized weight with the compensation of this low-rank approximation. This process alternates until a predetermined number of iterations is reached or a convergence condition is satisfied for the following optimization problem:
\begin{equation}\label{eq:fine-tune-optimal}
\min _{\mathbf{Q}, \mathbf{A}, \mathbf{B}}\left\|\mathbf{W}-Q(\mathbf{W})-\mathbf{A} \mathbf{B}^{\top}\right\|_F
\end{equation}
where $||.||_F$ is the Frobenius Norm of a matrix. Initialize $A$ and $B$ by minimizing Eq.~\ref{eq:fine-tune-optimal} will minimize the impact of the quantization operation over the weight, leading to a better quantization behavior. 

\begin{table*}[t]
\centering

\small
\begin{tabular}{cccccccc}
\toprule
Timesteps             & \begin{tabular}[c]{@{}c@{}}Bit-width (W/A)\end{tabular} & Method & Size (MB) & IS $\uparrow$ & FID $\downarrow$ & sFID $\downarrow$ & Precision $\uparrow$ \\ \midrule
\multirow{11}{*}{100} & \textcolor{gray}{32/32}                                   & \textcolor{gray}{FP}       & \textcolor{gray}{2575.42}   & \textcolor{gray}{274.78} &\textcolor{gray}{5.00} & \textcolor{gray}{19.02}  & \textcolor{gray}{0.8149} \\ \cmidrule{2-8} 
                      & \multirow{5}{*}{8/8}                                      & PTQ4DM   & 645.72   & 172.37 & 15.36 & 79.31 & 0.6926 \\
                      &                                                           & Q-Diffusion & 645.72 & 202.84 & 7.93 & 19.46 & 0.7299 \\
                      &                                                           & PTQD      & 645.72  & 199.00 & 8.12 & 19.64 & 0.7295 \\
                      &                                                           & RepQ*    & 645.72    & 254.70 & 5.20 & 19.87  & 0.7929 \\
                      &                                                           & PTQ4DiT \cellcolor{blue!10}  & 645.72 \cellcolor{blue!10}  & \textbf{277.27} \cellcolor{blue!10} & \textbf{4.73} \cellcolor{blue!10} & \textbf{17.83} \cellcolor{blue!10}  & \textbf{0.8270} \cellcolor{blue!10}\\
                      &                                                           & \textbf{Ours}  & 645.72     & 252.33 & 5.83 & 19.09  & 0.8032 \\ \cmidrule{2-8} 
                      & \multirow{5}{*}{4/8}                                      & PTQ4DM    & 323.79 & 26.02 & 89.78 & 57.20 & 0.2146 \\
                      &                                                           & Q-Diffusion & 323.79 & 42.80 & 54.95 & 36.13 & 0.3846 \\
                      &                                                           & PTQD      & 323.79 & 42.87 & 55.96 & 37.24 & 0.3948 \\
                      &                                                           & RepQ*     & 323.79 & 91.39  & 26.64 & 29.42 & 0.4347 \\
                      &                                                           & PTQ4DiT   & 323.79  & 190.38  & 7.75 & 22.01 & 0.7292 \\                      
                      &                                                           & \textbf{Ours}  \cellcolor{blue!10} & 323.79 \cellcolor{blue!10}  & \textbf{218.04} \cellcolor{blue!10}  & \textbf{6.86} \cellcolor{blue!10} & \textbf{19.64} \cellcolor{blue!10} & \textbf{0.7638} \cellcolor{blue!10}\\ \midrule 
\multirow{11}{*}{50}  & \textcolor{gray}{32/32}                                   & \textcolor{gray}{FP}     & \textcolor{gray}{2575.42}     &\textcolor{gray}{240.74}  &\textcolor{gray}{6.71}           &\textcolor{gray}{21.21}                 &\textcolor{gray}{0.7814} \\ \cmidrule{2-8} 
                      & \multirow{5}{*}{8/8}                                      & PTQ4DM & 645.72     & 154.08 & 17.52 & 84.28 & 0.6574 \\
                      &                                                           & Q-Diffusion & 645.72 & 153.01 & 14.61 & 27.57 & 0.6601 \\
                      &                                                           & PTQD    & 645.72    & 151.60 & 15.21 & 27.52 & 0.6578 \\
                      &                                                           & RepQ*  & 645.72      & 224.83 & 7.17 & 23.67 & 0.7496 \\
                      &                                                           & PTQ4DiT \cellcolor{blue!10}  & 645.72  \cellcolor{blue!10}   & \textbf{250.68} \cellcolor{blue!10} & \textbf{5.45} \cellcolor{blue!10} & \textbf{19.50} \cellcolor{blue!10} &  \textbf{0.7882} \cellcolor{blue!10}\\                      
                      &                                                           & \textbf{Ours}    & 645.72    &223.83 & 7.61 & 21.77 &  0.7654 \\ \cmidrule{2-8} 
                      & \multirow{5}{*}{4/8}                                      & PTQ4DM   & 323.79  & 19.29 & 102.52 & 58.66 & 0.1710 \\
                      &                                                           & Q-Diffusion & 323.79 & 109.22 & 22.89 & 29.49 & 0.5752 \\
                      &                                                           & PTQD   & 323.79    & 104.28 & 25.62 & 29.77 & 0.5667 \\
                      &                                                           & RepQ*     & 323.79  & 80.64 & 31.39 & 30.77 & 0.4091 \\
                      &                                                           & PTQ4DiT   & 323.79   & 179.95 & 9.17 & 24.29 & 0.7052 \\
                      &                                                           & \textbf{Ours} \cellcolor{blue!10}   & 323.79 \cellcolor{blue!10}  & \textbf{194.34} \cellcolor{blue!10} & \textbf{9.05} \cellcolor{blue!10} & \textbf{22.56} \cellcolor{blue!10} & \textbf{0.7285} \cellcolor{blue!10}\\ \bottomrule
\end{tabular}
\caption{Performance comparison on ImageNet 256$\times$256 (cfg=1.50). `(W/A)' indicates that the precision of weights and activations are W and A bits, respectively.}
\label{tab:256_comparison}
\end{table*}

\section{Experiments}
In this section, we evaluate the performance of DiTAS in generating conditional images on ImageNet at different resolutions, including $256\times 256$ and $512 \times 512$. To evaluate the effectiveness of the DiTAS algorithm, we also conducted sampling tests across different bit-widths. To better validate the effectiveness of DiTAS, we set up a baseline called ~\textbf{LinearQuant}, which applies the Asymmetric Post-Training Quantization mentioned earlier in Eq.~\ref{eq:ptq} to quantize both the DiT weights and activations. We evaluate the performance of DiTAS by comparing it with Q-Diffusion~\cite{li2023q}, PTQ4DM~\cite{ptq4dm}, PTQD~\cite{he2024ptqd}, and RepQ*~\cite{li2023repq}. Furthermore, we compare the PTQ4DiT~\cite{wu2024ptq4dit}, a recently poposed DiT quantization framework that previously achieves the state-of-the-arts performance. Both our method and the mentioned approaches use DDPM as the sampler, providing a consistent condition for comparison. 
However, Q-DiT~\cite{chen2024q}, although a quantization method for DiT, uses DDIM as its sampler; therefore, we do not include it in the comparison. Moreover, we conduct ablation experiments to show DiTAS performance under different hyperparameter settings. 

In Section~\ref{sec:4.1}, we outline the experimental setting details. Performance comparisons of quantized diffusion transformers can be found in Section~\ref{sec:4.2}. Sections~\ref{sec:4.3} and \ref{sec:4.4} conduct evaluations of conditional generation on ImageNet at resolutions of $256\times 256$ and $512\times512$, respectively. The results of the ablation study, evaluating each component of DiTAS, are presented in Section~\ref{sec:4.5}.

\subsection{Experiment Settings}
\label{sec:4.1}
\paragraph{Models and metrics}
We download the pretrained DiT model from the official Huggingface website~\cite{huggingface} and apply the quantization techniques described in Section~\ref{sec:methodology} over it. We evaluate DiTAS on the ImageNet dataset~\cite{deng2009imagenet} with two different sizes of generated images: \(256 \times 256\) and \(512 \times 512\). The quality of the generated images is evaluated using metrics including Inception Score (IS), Fréchet Inception Distance (FID)~\cite{heusel2017gans}. Results are obtained by sampling $10,000$ images for ImageNet $256\times256$ and $5,000$ images for ImageNet $512\times512$ as previous works~\cite{nichol2021improved, ptq4dm}. We evaluate them with ADM’s TensorFlow evaluation suite~\cite{dhariwal2021diffusion}. The evaluation is performed on a single A100 GPU. To compute the TAS factor, 12 samples are selected as calibration dataset, each of which is generated from a different class randomly chosen from 1000 classes on ImageNet. When conducting grid search optimization, the same 12 samples are used for calibration purpose. The search space for \(\alpha\) spans the interval \([0, 1]\). The LoRA integrated into the weights has a rank of 32, and the AO is performed for 10 iterations. For evaluation, the setting is the same as the original DiT settings described in~\cite{peebles2023scalable}.

\begin{table}[t]
\centering
\setlength{\tabcolsep}{3pt}
\small
\begin{tabular}{cccccc}
\toprule
Timesteps            & Method     & FID $\downarrow$ & sFID $\downarrow$ & IS $\uparrow$ & Precision $\uparrow$ \\ \midrule
\multirow{6}{*}{100} & \textcolor{gray}{FP}          & \textcolor{gray}{9.06} & \textcolor{gray}{37.58} & \textcolor{gray}{239.03} & \textcolor{gray}{0.8300}  \\ \cmidrule{2-6} 
                     & PTQ4DM     & 70.63            & 57.73             & 33.82         & 0.4574 \\
                     & Q-Diffusion & 62.05            & 57.02             & 29.52         & 0.4786 \\
                     & PTQD       & 81.17            & 66.58             & 35.67         & 0.5166 \\
                     & RepQ*      & 62.70            & 73.29             & 31.44         & 0.3606 \\
                    & PTQ4DiT      & 19.00   & 50.71            &121.35 & 0.7514 \\
                     \rowcolor{blue!10} & \textbf{Ours}       & \textbf{13.45}   & \textbf{40.92}    & \textbf{183.36} & \textbf{0.7986} \\ \midrule
\multirow{6}{*}{50}  & \textcolor{gray}{FP}         & \textcolor{gray}{11.28} & \textcolor{gray}{41.70} & \textcolor{gray}{213.86} & \textcolor{gray}{0.8100}  \\ \cmidrule{2-6} 
                     & PTQ4DM     & 71.69            & 59.10             & 33.77         & 0.4604 \\
                     & Q-Diffusion & 53.49            & 50.27 & 38.99 & 0.5430 \\
                     & PTQD       & 73.45            & 59.14             & 39.63         & 0.5508 \\
                     & RepQ*      & 65.92            & 74.19             & 30.92         & 0.3542 \\
                     & PTQ4DiT      & 19.71   & 52.27             &118.32 & 0.7336 \\
                     \rowcolor{blue!10} & \textbf{Ours}       & \textbf{17.92}   & \textbf{45.28}             & \textbf{147.08} & \textbf{0.7612} \\ \bottomrule
\end{tabular}
\caption{\small Performance on ImageNet 512$\times$512 with W4A8 (cfg=1.50).}
\label{tab:512_performance}
\end{table}

\begin{table*}[h]
\centering
\begin{tabular}{cccccc}
\toprule
Method                    & \begin{tabular}[c]{@{}c@{}}Bit-width (W/A)\end{tabular} & FID$\downarrow$     & sFID$\downarrow$   & IS$\uparrow$ & \begin{tabular}[c]{@{}c@{}}Precision$\uparrow$ \end{tabular}     \\ \midrule
FP32                                &\textcolor{gray}{32/32}          &\textcolor{gray}{6.71}           &\textcolor{gray}{21.21}         &\textcolor{gray}{240.74}         &\textcolor{gray}{0.7814}         \\ \midrule
LinearQuant (Baseline)                               & 4/8          &128.76             &65.81       &12.09        &0.1030         \\
+ Temporal-aggregated Smoothing (TAS)                     &4/8      &22.31        &32.38      &115.33     &0.5743          \\
+ Grid Search Optimization    &4/8       &17.65          &31.51      &138.72     & 0.6378\\
\rowcolor{blue!10}+ Advanced Weight Quantization             &4/8   & \textbf{9.05} & \textbf{22.56} & \textbf{194.34} & \textbf{0.7285} \\ \bottomrule
\end{tabular}
\caption{\small The effect of different components proposed in the paper. The experiment is conducted under W4A8 configuration over DiT-XL/2 on ImageNet $256 \times 256$ with cfg=1.5 and time steps are 50.}
\label{tab:ablation}
\end{table*}

\paragraph{Implementation of Quantization}

We use per-input-channel asymmetric quantization for weights and per-tensor dynamic asymmetric quantization for activations, following the standard approach adopted by previous works. DiTAS is evaluated over different bit-widths of activations and weights including W8A8 and W4A8. In the specific instance of quantized matrix multiplication, we employ a per-input-channel weight quantization, which necessitates an outer product approach. Additionally, since the smoothing factor has the same dimensionality as the scaling factor used in weight quantization, integrating the smoothing factor into the scaling factor in the de-quantization process can improve computational efficiency and avoid increasing the complexity of weight quantization. Therefore, in our enhanced weight quantization scheme, we combine the smoothing factor with the scaling factor. 

We quantize all DiT modules to their target bit-width settings, except for the conditioning MLP. Due to its minimal computational overhead, we retain the conditioning MLP at FP32 precision to ensure accuracy, a practice also adopted by other DiT quantization methods~\cite{chen2024q, wu2024ptq4dit}.
\begin{figure}[t]
  \centering
   \includegraphics[width=0.99\linewidth, trim={0cm 0cm 0cm 0cm}, clip]{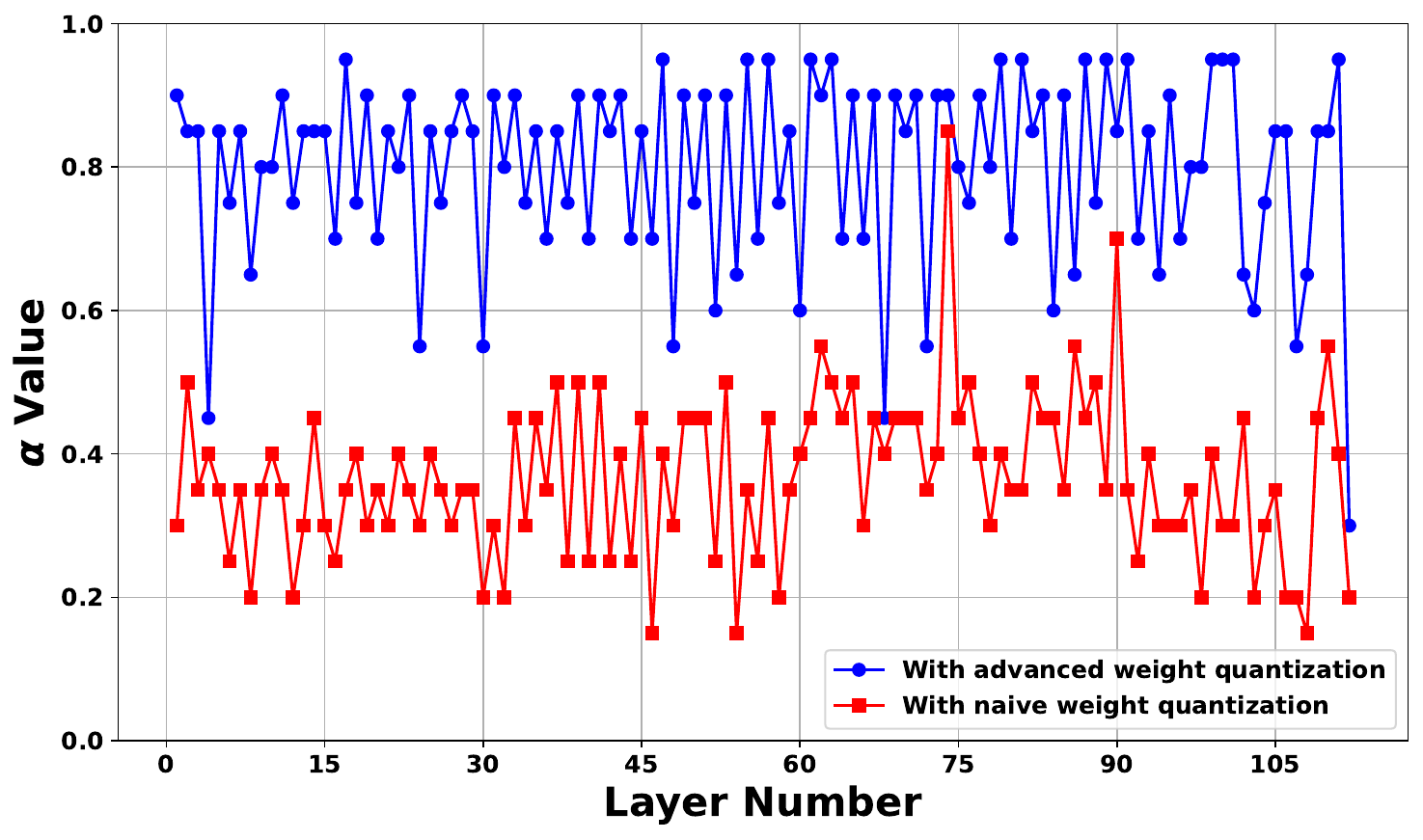}
   \caption{Grid search comparison under W4A8 configuration on ImageNet $256 \times 256$ (cfg=1.5, 50 steps).}
   \label{fig:search}
\end{figure}

\subsection{Performance Evaluation}
\label{sec:4.2}
In this section, we conduct comparison between the DiTAS and other quantization methods. Figures~\ref{W8A8} and \ref{W4A8} depict the sample images generated by DiTAS under two different precision settings: W8A8 and W4A8. It is clear that DiTAS can produce images with quality comparable to photorealistic images across various bit-width settings. When the bit-widths are set to W4A8, our approach achieves FID values of $6.86$ with sampling 100 steps and $9.05$ for 50 sampling steps, respectively. In contrast, PTQ4DiT, the previous state-of-the-art, achieves FID values of 10.05 and 8.74 under the same configurations. Our approach outperforms this, demonstrating superior performance.

\subsection{Evaluation with ImageNet $256\times256$}
\label{sec:4.3}
To evaluate conditional generation on ImageNet $256\times256$, experiments are conducted on the performance of various quantization techniques under different bit-width and time step configurations. PTQ4DiT is the previous state-of-the-art DiT quantization method. As shown in Table~\ref{tab:256_comparison}, DiTAS outperforms PTQ4DiT, achieving new state-of-the-art results in the 8-bit activation with 4-bit weight configuration (W4A8) across various timestep settings. Specifically, with 100 timesteps and under W4A8 configuration, DiTAS achieves an impressive 218.04 in IS, outperforming PTQ4DiT's 190.38, while also maintaining a lower FID of 6.86 compared to PTQ4DiT's 7.75. This indicates that DiTAS not only generates higher quality images but also achieves this with greater consistency. The sFID metric further supports this, with DiTAS scoring 19.64 versus PTQ4DiT's 22.01, suggesting that DiTAS produces images that are closer to the ground truth distribution. At 50 timesteps, DiTAS continues to show its prowess, achieving an IS of 194.34 compared to PTQ4DiT's 179.95, and a slightly lower FID of 9.05 versus PTQ4DiT's 9.17. The sFID scores are also indicative of DiTAS's superior performance, with a score of 22.56 for DiTAS and 24.29 for PTQ4DiT.

It is noteworthy that for W8A8, DiTAS slightly underperforms PTQ4DiT. This is attributed to our focus on optimal activation smoothing, which inherently considers the adverse impact of transferring quantization to weights. Our approach, when combined with 4-bit weights, leverages the joint quantization of weights and activations to showcase greater benefits, as evidenced by the superior results in the W4A8 configuration.

In summary, DiTAS establishes itself as a leading contender in the realm of low-bit quantization, particularly excelling in the W4A8 scenario, where it consistently outperforms PTQ4DiT across different time steps' setting, showcasing its potential for real-world applications requiring efficient and high-quality image generation.

\subsection{Evaluation over ImageNet $512\times512$}
\label{sec:4.4}

The evaluation results on ImageNet $512\times512$ is shown in table~\ref{tab:512_performance}. We observe that DiTAS significantly outperforms other quantization techniques in the W4A8 configuration across various timestep settings, nearly matching the performance of the full precision model. Specifically, with 100 timesteps, our method achieves an FID of 13.45 and an IS of 183.36, which shows 5.55 FID reduction and 62.01 IS improvement of the previous tate-of-the-art PTQ4DiT. Furthermore, we also evaluate the performance with 50 timesteps, as shown in the bottom half of the table. Again, DiTAS demonstrates the best performance, with an FID of 17.92 and an IS of 147.08, showing our method's robustness across different conditions. In conclusion, the experimental results validate the effectiveness of DiTAS for quantizing DiT with low precision.

\subsection{Ablation Study}
\label{sec:4.5}
In this section, we investigate the individual impact of each method proposed in Section~\ref{sec:methodology} on the DiTAS performance, with the results presented in Table~\ref{tab:ablation}. All experiments are conducted using DiT-XL/2 on ImageNet at \(256 \times 256\) resolution, with cfg set to 1.5 and timesteps are set to 50. Table~\ref{tab:ablation} shows the individual impact of each methods described in Section~\ref{sec:methodology} over the DiTAS performance.

Starting with the baseline LinearQuant method, we note a marked decrease in performance when quantizing to 4 bits for weights and 8 bits for activations, resulting in a substantial increase in FID and a decrease in IS and Precision. However, by integrating our proposed temporal-aggregated smoothing (TAS), there is a significant improvement, slashing the FID to 22.31 and boosting the IS and Precision scores considerably. Further enhancements are achieved with the addition of Grid Search Optimization, fine-tuning the TAS factor. This step significantly leads to a more refined FID of 17.65 and IS of 138.72.  Lastly, the introduction of Advanced Weight Quantization can reduce the difficulty of weight quantization and helps DiTAS to achieve the cutting-edge performance, with an FID of 9.05, an sFID of 22.56, and an IS score of 194.34, closely rivaling the unabridged FP32 model. 
\paragraph{Visualization}
The results from Figure~\ref{fig:search} show that the alpha values obtained from the grid search without advanced weight quantization are consistently lower across different layers compared to those obtained with advanced weight quantization. This demonstrates that by addressing the challenge of quantizing weights, the grid search can more effectively reduce activation outliers, resulting in a significant performance improvement. On the contrary, when using naive weight quantization, it is crucial to prevent too much of the activation's difficulty from being shifted onto weight quantization, as this would lead to an overall decline in performance. In summary, advanced weight quantization allows TAS and grid search optimization to more effectively enhance activation quantization performance.

\section{Conclusion}
In this paper, we introduce DiTAS, an efficient data-free Post-Training Quantization method tailored for low-precision DiT execution. To tackle the challenges associated with quantizing activations, we employ temporal-aggregated smoothing (TAS) techniques to eliminate outliers. Additionally, we introduce the grid search to better optimize the TAS factor. Experimental results demonstrate that our approach enables 4-bit weight, 8-bit activation (W4A8) quantization for DiTs while maintaining comparable performance to the full-precision model.

\newpage

{\small
\bibliographystyle{ieee_fullname}
\bibliography{egbib}
}

\end{document}